\documentclass{article}

\usepackage[preprint]{corl_2026} 

\usepackage[utf8]{inputenc}
\usepackage{booktabs}
\usepackage{tabularx}
\usepackage{geometry}
\usepackage{amsfonts}
\usepackage{amsmath}
\usepackage[table]{xcolor} 
\usepackage{graphicx}
\usepackage{caption}
\usepackage{wrapfig}
\usepackage{float}
\usepackage{threeparttable}
\usepackage{caption}
\usepackage{enumitem}
\usepackage{subcaption}
\usepackage{multirow}
\usepackage{threeparttable}
\usepackage{algorithm}
\usepackage{algpseudocode}
\usepackage{titletoc}

\graphicspath{{images/}}

\hypersetup{pdftitle={Keeping Jerk in Check: Jointly Trained Perception-Gated Experts for Smooth Terrain Transitions},
  pdfauthor={Francisco Affonso, Matheus P. Angarola, Ana Luiza Mineiro, Aditya Potnis, Marcelo Becker, Girish Chowdhary},
  pdfsubject={}}

\newcommand{\hd}[2]{\textbf{\begin{tabular}[b]{@{}c@{}}#1\\#2\end{tabular}}}

\newcolumntype{L}{>{\raggedright\arraybackslash}X}
\newcolumntype{Y}{>{\centering\arraybackslash}X}
\newcommand{\cmark}{\textcolor{green!40!black}{\checkmark}}

\title{Keeping Jerk in Check: \\ Jointly Trained Perception-Gated Experts for Smooth Terrain Transitions}

\author{
  Francisco Affonso$^1$, Matheus P. Angarola$^2$, Ana Luiza Mineiro$^2$, Aditya Potnis$^1$,\\ \textbf{Marcelo Becker}$^2$, \textbf{Girish Chowdhary}$^1$\\
    $^1$ University of Illinois Urbana-Champaign \quad $^2$ University of São Paulo\\
}

\begin{document}
\maketitle

\renewcommand{\thefootnote}{\fnsymbol{footnote}}
\footnotetext{Correspondence to \href{mailto:faffonso@illinois.edu}{\fontfamily{pcr}\selectfont faffonso@illinois.edu}}

\renewcommand{\thefootnote}{\fnsymbol{footnote}}
\setcounter{footnote}{1}

\vspace{-2em}
\begin{figure}[H]
    \centering
    \includegraphics[width=\textwidth]{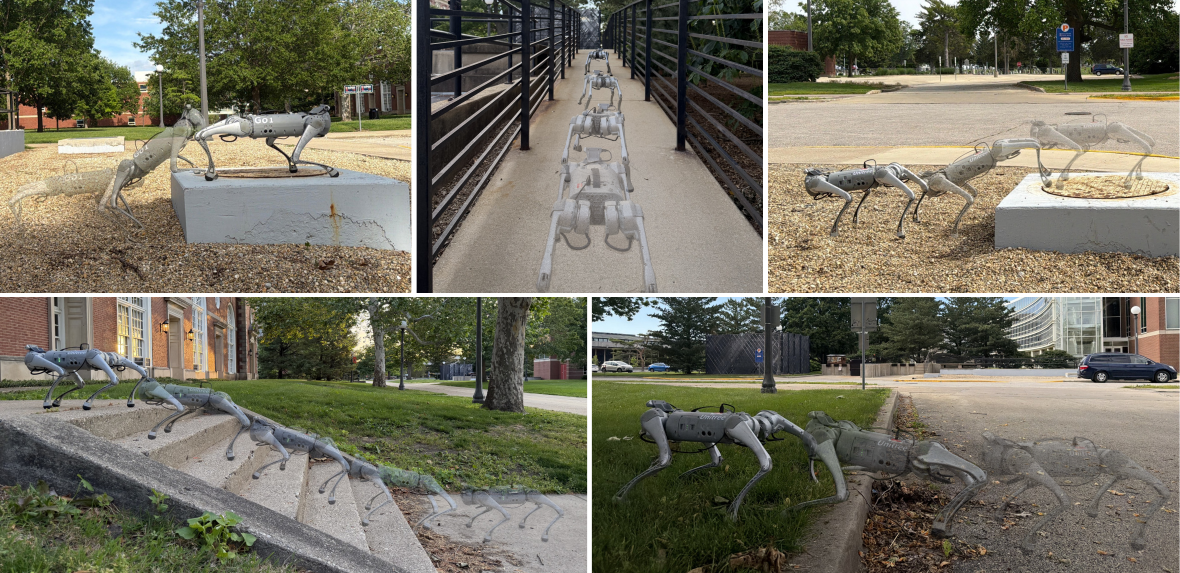}
    \caption{Real-world deployment across discontinuous outdoor terrain. Perception-conditioned expert composition enables adaptation without terrain labels or high-level policy selection.
}
    \label{fig:teaser}
\end{figure}
\vspace{-2em}


\begin{abstract}
  Perceptive legged locomotion across discontinuous terrain requires shared locomotion knowledge, task-specific maneuvers, and smooth transitions between them. Existing approaches train specialized policies independently and either switch among them with a hierarchical selector or distill them into a monolithic policy. Both fix the constituent behaviors before they are composed, and both are assessed through aggregate course outcomes that leave behavior at terrain boundaries uncharacterized. We instead propose CTS-MoE, which formulates multi-terrain locomotion as multi-task reinforcement learning and jointly trains a perception-gated dense mixture-of-experts actor with task-specific critics: soft routing composes expert knowledge into continuously varying behavior, while separated critics isolate heterogeneous objectives so that specialization survives joint optimization. A concurrent teacher--student formulation transfers this composition from privileged terrain information to onboard depth, so that at deployment routing is inferred from perception alone, with no terrain label, selector, or distillation stage. Against hierarchical and DAgger-distilled baselines, the policy raises connected-course success to 93.3\% from 73.3\% and 53.3\%, lowers mean action jerk by about a third across the course, and avoids the switching transient that hard selection introduces at boundaries. On a Unitree Go1, it completes an extended outdoor course in 10/10 trials and transfers to four outdoor terrain types. Project Website: \url{https://cts-moe.github.io/}
\end{abstract}


\section{Introduction}

Reinforcement learning (RL) has enabled robust perceptive locomotion by implicitly incorporating complex physical phenomena directly in the policy and reasoning about terrain geometry through perceptual input~\cite{miki2022learning,hoeller2024anymal,cheng2024extreme,wu2025vocaloco}. When deployed, these policies face multiple terrains where the robot must reuse shared gait structure, execute terrain-specific maneuvers, and recombine them at boundaries whose particular pairings may never occur during training. Multi-terrain locomotion is therefore a compositional generalization problem, yet existing methods learn per-terrain behaviors and report aggregate completion, leaving controller behavior at terrain boundaries, where anticipatory behavior must change, largely uncharacterized.

Two difficulties stand in the way. First, different skills call for different objectives. Velocity tracking suits continuous locomotion, whereas gaps and climbable obstacles benefit from waypoint objectives that permit transient acceleration, deceleration, and pausing~\cite{rudin2022advanced,chamorro2024stair}. Optimized jointly through one value function, these heterogeneous returns interfere, suppressing specialization or biasing learning toward the easiest task~\cite{mysore2022multi}. Second, the skills must be composed into one deployable controller. Hierarchical and distillation approaches both optimize a set of specialist policies separately, one per terrain, and then either switch among them~\cite{hoeller2024anymal,wu2025vocaloco} or merge them into a single policy~\cite{zhuang2023robot}; in both, the constituent behaviors are fixed before they are composed.

We instead train specialization and composition jointly in CTS-MoE, a Concurrent Teacher–Student Mixture-of-Experts framework. Task-specific rewards induce distinct traversal behaviors, task-specific critics let those behaviors survive joint optimization, and a dense mixture-of-experts (MoE) actor, routed from perception, composes them continuously. Concurrent teacher--student training~\cite{wang2024cts} transfers the routing from privileged terrain information to onboard depth, so at deployment the composition is inferred from perception alone. Our contributions are:
\begin{itemize}[nosep,leftmargin=*]
    \item An asymmetric multi-task formulation in which task-specific rewards and critics isolate heterogeneous returns, so a single shared actor learns specialized behaviors jointly.
    \item A perception-conditioned MoE router, transferred to onboard depth, that composes these behaviors without a terrain label, selector, or distillation stage.
    \item A transition-centric evaluation against hierarchical and distilled baselines retrained under a matched protocol, measuring success on a connected course with held-out segments and switch-aligned action jerk, followed by hardware deployment on a Unitree Go1.
\end{itemize}
\section{Related Work}

\begin{wraptable}{r}{0.47\textwidth}
    \centering
    \vspace{-13pt}
    \small
    \resizebox{\linewidth}{!}{%
    \begin{tabular}{@{}lcccc@{}}
        \toprule
        \textbf{Method} & \textbf{Perc.} & \textbf{MoE} & \textbf{Multi-Task} & \textbf{J-Train} \\ \midrule
        CTS~\cite{wang2024cts} & & & & \cmark \\
        Ego-Vision~\cite{agarwal2023legged} & \cmark & & & \\
        Parkour Learning~\cite{zhuang2023robot} & \cmark & & \cmark & \\
        VOCALoco~\cite{wu2025vocaloco} & \cmark & & \cmark & \\
        MoE-Loco~\cite{huang2025moe} & & \cmark & \cmark & \cmark \\ \midrule
        \textbf{CTS-MoE (Ours)} & \cmark & \cmark & \cmark & \cmark \\ \bottomrule
    \end{tabular}%
    }
    \caption{Comparison of frameworks. \textit{Perc.} denotes depth-based
    perception. \textit{J-Train} indicates end-to-end optimization,
    avoiding sequential distillation or high-level selection.}
    \label{tab:comparison}
    \vspace{-12pt}
\end{wraptable}

\textbf{Teacher--student learning.} Privileged teachers transfer terrain and dynamics information to students acting from deployable observations~\cite{kumar2021rma,agarwal2023legged}. CTS~\cite{wang2024cts} trains both concurrently, reducing the mismatch of sequential distillation (see App.~\ref{app:cts}). These methods address representation transfer, not how one actor should compose behaviors induced by heterogeneous objectives.

\textbf{Skill specialization and composition.} Hierarchical approaches train separate specialists and activate them with a high-level selector~\cite{hoeller2024anymal,wu2025vocaloco,angarola2025learning}, so deployment depends on predefined skill categories, accurate switching, and coverage of boundary states. Distilling specialists with DAgger~\cite{ross2011reduction,zhuang2023robot} removes switching, but the teachers never co-adapt and the result depends on the transition coverage of the collected demonstrations.

\textbf{Mixture-of-experts and multi-critic RL.} MoE architectures jointly train subnetworks combined by learned routing, capturing diverse motions in humanoid control~\cite{chen2025gmt} and manipulation~\cite{hao2026abstracting}; multi-critic methods reduce interference with per-task value heads~\cite{mysore2022multi} but require an explicit task label.

Closest to ours, MoE-Loco~\cite{huang2025moe} routes a multi-task locomotion MoE from proprioception alone, which can react to terrain only after contact. Discontinuous terrain instead requires changing the expert composition before the boundary; we therefore route from perception and pair it with task-specific critics in a concurrent teacher--student framework (Table~\ref{tab:comparison}).
\section{Problem Formulation}

We formulate perceptive multi-terrain locomotion as a partially observable Markov decision process (POMDP). The policy does not directly observe the complete state of the robot and its environment. An encoder maps the observation history to a latent representation \(z_t\in\mathcal{Z}\), from which a policy \(\pi(a_t\mid z_t)\)\footnote{For simplicity, we assume that the encoded observation history forms a valid Markovian state. Although this assumption does not hold in general, it is a common practical approximation.} then selects an action \(a_t\in\mathcal{A}\) from this latent representation.

We further cast the training terrains as \(N\) related tasks indexed by \(c_t\in\{1,\ldots,N\}\). These tasks share the robot dynamics and common regularization objectives but require distinct traversal behaviors. The environment evolves according to the latent transition distribution \(p(s_{t+1}\mid s_t,a_t)\), observations are drawn from an emission model \(o_t\sim p(\cdot\mid s_t)\), and each task defines a specialized reward \(r^{(c_t)}(s_t,a_t)\). The policy is trained to maximize:
\begin{equation}
J(\pi)=\mathbb{E}_{\tau\sim\pi}\left[
\sum_{t=0}^{\infty}\gamma^t r^{(c_t)}(s_t,a_t)
\right],
\label{eq}
\end{equation}
where \(\tau\) denotes an on-policy trajectory and \(\gamma\in[0,1)\) is the discount factor.

\textbf{Observations.} The student receives only a deployable history $o_{t-H:t}$ of proprioception $o^p$ and depth images $d_t$. During training, the teacher additionally receives a privileged state $s_t$ comprising a ground-truth heightmap $h_t$, environment parameters $e_t$, and the task index $c_t$. Separate encoders map each group to a latent of equal dimension,
\begin{equation}
z_t^s=\psi_{\theta_s}^s(o_{t-H:t}), \qquad
z_t^t=\psi_{\theta_t}^t(s_t),
\label{eq:latent_states}
\end{equation}
and only $z_t^s$ is available at deployment. App.~\ref{app:observations} details every observation term and App.~\ref{app:encoders} the encoder architectures. The action \(a_t\in\mathbb{R}^{12}\) specifies target joint positions tracked by low-level PD controllers.

\section{Method}

\begin{figure}[t]
    \centering
    \includegraphics[width=0.95\linewidth]{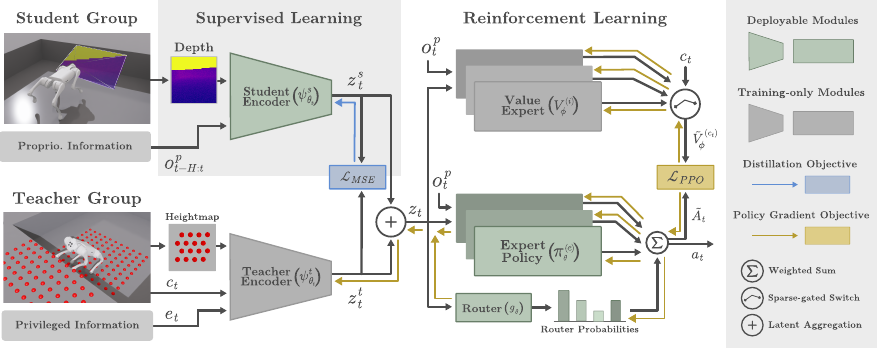}
    \caption{Overview of CTS-MoE. Teacher and student groups are trained concurrently. Privileged information is encoded by the teacher, while the teacher latent is regressed by the student encoder from deployable observations. A shared MoE actor, whose router composes the experts, is driven by both groups, and per-task values are provided by task-specific critics to stabilize PPO.}
    \label{fig:main_diagram}
    \vspace{-8pt}
\end{figure}

Two competing requirements are resolved by CTS-MoE (Fig.~\ref{fig:main_diagram}). Objectives that should not share value estimates are separated by task-specific rewards and critics, while behaviors with common locomotion structure are shared and composed by a dense MoE actor, and the routing representation is transferred to onboard perception through concurrent teacher--student training.

\textbf{Task-Specific Reward Design.} Maneuvers useful only on particular terrain can be suppressed by a single reward. Following prior designs~\cite{wu2025vocaloco,rudin2022learning,wu2026robogauge}, each task reward is composed of task objectives, shared regularizers, style terms, and termination penalties, and only the terms that induce different traversal behavior are varied by task. Velocity tracking is used on flat ground and stairs, since their motion is continuous. Gaps and climbable obstacles require transient acceleration, deceleration, or pausing, which would be penalized by a fixed-speed target. These tasks are instead rewarded for waypoint proximity along the commanded direction and through a velocity kernel centered on the commanded speed, so that average motion is tied to the command while brief deviations are tolerated. The reward is selected by the training-only index $c_t$ (see App.~\ref{app:simulation}, Table~\ref{tab:rewards}).

\textbf{Task-Specific Critics and Normalization.} Heterogeneous return distributions are created by these rewards, and with a shared critic, incompatible targets would be fitted while the advantages were dominated by high-magnitude tasks. 

Each task is therefore assigned a value head $V_\phi^{(k)}$, updated only from transitions generated under $r^{(k)}$. Since scale is not equalized by separation alone, PopArt~\cite{van2016learning} is adopted, and each task's return mean and variance are tracked by an exponential moving average. With $\widetilde{R}_t^{(k)}$ and $\widetilde{V}_\phi^{(k)}$ denoting the normalized return and prediction, the critic is trained to minimize
\begin{equation}
  \mathcal{L}_{\text{value}}(\phi) = \frac{1}{N}\sum_{k=1}^{N}
  \mathbb{E}_{\mathcal{D}^{(k)}}\Big[\mathcal{H}_{\delta}\big(
  \widetilde{V}^{(k)}_{\phi}(z_t \oplus o^p_t), \widetilde{R}^{(k)}_t\big)\Big],
  \label{eq:value}
\end{equation}
where $\mathcal{D}^{(k)}$ holds the transitions of task $k$ and $\mathcal{H}_\delta$ is the Huber loss. The output layer is rescaled by PopArt whenever its statistics change, and advantages are normalized per task before the policy update.

\textbf{Perception-Routed MoE Actor.} Specialization must be retained by the actor without a task label at deployment. Soft expert weights are therefore inferred from the latent by a router, and the $E$ expert outputs are composed in action space by a dense mixture,
\begin{equation}
  \pi_{\theta}(a_t \mid z_t, o^{p}_t) = \mathcal{N}\big(\bar{\mu}_t,
  \sigma^{2}\big), \qquad
  \bar{\mu}_t = \sum_{e=1}^{E} g_e(z_t)\, \mu_e(z_t, o^{p}_t),
  \label{eq:moe}
\end{equation}
where $\mu_e$ is the mean action of expert $e$, $g_e$ its softmax routing weight, and $\sigma$ a state-independent learned standard deviation. 

Because means rather than densities are composed, the policy remains Gaussian and the PPO probability ratio stays exact. The same actor is fed $z^{t}_t$ by the teacher group and $z^{s}_t$ by the student group. Since $c_t$ is included in $z^{t}_t$, teacher routing is partly label-driven during training, whereas the same routing must be reproduced by the student from depth alone. Premature collapse onto a single expert is discouraged by router entropy, load-balancing, and logit-magnitude regularizers (see App.~\ref{app:objectives}).

\textbf{Concurrent Teacher--Student Optimization.} Following CTS~\cite{wang2024cts}, the deployable representation is learned concurrently rather than by distilling a fixed policy in a second stage (App.~\ref{app:cts_background}). Datasets $\mathcal{D}_t$ and $\mathcal{D}_s$ are collected by parallel teacher and student groups, and the clipped PPO surrogate is maximized on each with per-task normalized advantages. These updates are applied to the actor and teacher encoder together with Eq.~\ref{eq:value} and the router regularizers. After each PPO update, only the student encoder is supervised with student trajectories, by minimizing
\begin{equation}
  \mathcal{L}_{\text{MSE}}(\theta_s) = \frac{1}{|\mathcal{D}_s|}
  \sum_{i \in \mathcal{D}_s} \big\lVert \psi^{s}_{\theta_s}(o_i) -
  \operatorname{sg}\big(\psi^{t}_{\theta_t}(s_i)\big) \big\rVert_2^{2}.
  \label{eq:distill}
\end{equation}
Because the student encoder is shielded from RL gradients, it is kept on the teacher target, while the shared actor, trained continually on student rollouts, co-adapts to the evolving student representation without a separate distillation stage.
\section{Experimental Results}
\label{sec:experiments}

All policies are trained in Isaac Lab~\cite{mittal2025isaac} with RSL-RL~\cite{schwarke2025rslrl} for 15,000 iterations and deployed on a Unitree Go1. Each of the six training terrains has six curriculum levels, while rough, discrete, and sloped terrain are withheld (see App.~\ref{app:simulation}). The reward and critic design is ablated in App.~\ref{app:reward_critic}.

\textbf{Baselines.} Since published results confound architecture with rewards, observations, terrains, and compute, each method's defining architecture is retrained with our rewards and common protocol.
\begin{itemize}[nosep, leftmargin=*]
    \item \textbf{Blind-Single.} Proprioception only, with a single actor~\cite{wang2024cts}.
    \item \textbf{Blind-MoE.} Proprioception only, with our MoE actor~\cite{huang2025moe}.
    \item \textbf{Perceptive-Single.} Depth and proprioception, with a single actor~\cite{agarwal2023legged}.
    \item \textbf{Hierarchical (Oracle).} Terrain specialists selected by $c_t$ at training and inference~\cite{angarola2025learning,wu2025vocaloco}. On unseen terrain, which has no specialist, the flat specialist is used, as it performed best there.
    \item \textbf{Distilled.} The same specialists distilled into one policy with DAgger~\cite{ross2011reduction}, following~\cite{zhuang2023robot}.
\end{itemize}

\begin{wraptable}{r}{0.54\textwidth}
    \centering
    \vspace{-12pt}
    \footnotesize
    \setlength{\tabcolsep}{2pt}
    \renewcommand{\arraystretch}{1.02}
    \caption{Success rate (\%) by terrain over 45 episodes. $^*$Unseen during training.}
    \label{tab:success_rate}
    \vspace{-6pt}
    \begin{tabular*}{\linewidth}{@{\extracolsep{\fill}}l*{6}{c}@{}}
        \toprule
        \textbf{Terrain} & \hd{}{Ours} & \hd{Perc.}{Single} & \hd{Blind}{MoE} & \hd{Blind}{Single} & \hd{Hier.}{(Oracle)} & \hd{}{Dist.} \\
        \midrule
        Flat      & \textbf{100.0} & \textbf{100.0} & \textbf{100.0} & 97.8 & \textbf{100.0} & \textbf{100.0} \\
        Ascend      & 95.6 & 91.1 & 71.1 & 84.4 & \textbf{100.0} & 93.3 \\
        Descend      & \textbf{100.0} & \textbf{100.0} & 97.8 & \textbf{100.0} & \textbf{100.0} & \textbf{100.0} \\
        Cross Gap & \textbf{97.8} & 68.9 & 40.0 & 40.0 & \textbf{97.8} & 93.3 \\
        Climb Up     & \textbf{86.7} & 75.6 & 62.2 & 77.8 & 80.0 & 64.4 \\
        Climb Down     & \textbf{97.8} & 95.6 & 95.6 & \textbf{97.8} & 84.4 & 84.4 \\
        Discrete$^*$ & \textbf{73.3} & 37.8 & 40.0 & 53.3 & 20.0 & 11.1 \\
        Rough$^*$  & \textbf{100.0} & \textbf{100.0} & \textbf{100.0} & \textbf{100.0} & \textbf{100.0} & \textbf{100.0} \\
        \bottomrule
    \end{tabular*}
    \vspace{-10pt}
\end{wraptable}

\textbf{Per-Terrain Results.} We evaluate forward commands $v_x\in\{0.5,0.75,1.0\}\,\mathrm{m/s}$ with $v_y=\omega_z=0$ and report success rates in Table~\ref{tab:success_rate} (tracking errors in App.~\ref{app:tracking}). Among deployable policies, the best or tied success on every terrain is attained by ours, and all methods saturate on the held-out rough terrain. On unseen discrete terrain, our policy reaches 73.3\%, above 53.3\% for the best deployable baseline, 20.0\% for the hierarchical oracle, and 11.1\% for the distilled policy, which is consistent with better generalization through composition of the learned experts.

The four variants, drawn from architectures and observation configurations in the literature, reveal how perception influences the benefit of MoE. Without perception, the mixture is detrimental, with 77.8\% average success over the six training terrains for Blind-MoE versus 83.0\% for Blind-Single, whereas with perception it is beneficial, with 96.3\% versus 88.5\% for Perceptive-Single. The benefit is therefore realized once routing is conditioned on perception, not through added capacity. When the router outputs are replaced with uniform weights $g_e=1/E$ at inference, success drops to $2.2\%$, so the behavior is carried by the learned routing rather than by the ensemble average (see App.~\ref{app:experts}).

\begin{figure*}[b]
    \centering
    \includegraphics[width=0.92\textwidth]{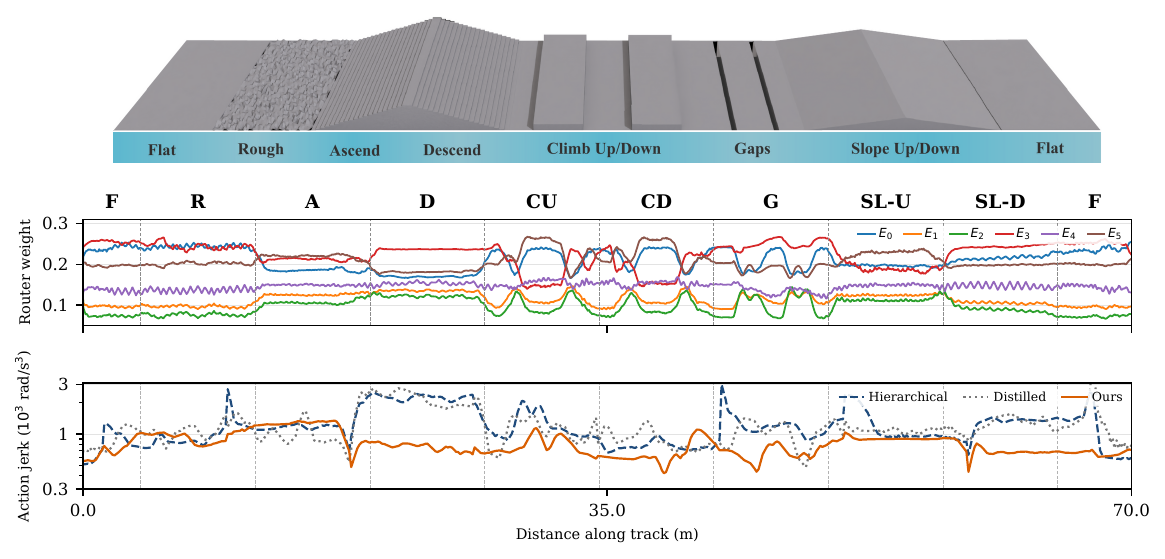}
    \vspace{-8pt}
    \caption{Connected-course evaluation. The multi-terrain map, the perception-conditioned router weights of our policy, and the action jerk along the same track are shown from top to bottom, with terrain boundaries aligned across the routing and jerk traces.}
    \label{fig:long_track_summary}
    \vspace{-2pt}
\end{figure*}

\begin{figure}[t]
    \centering
    \begin{subfigure}[b]{0.42\textwidth}
        \centering
        \includegraphics[width=\linewidth]{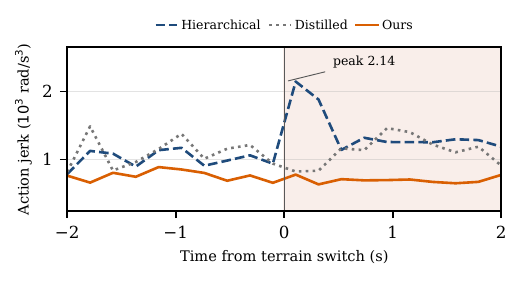}
        \vspace{-14pt}
        \caption{Simulation, aligned to terrain switches.}
        \label{fig:switch_aligned_jerk}
    \end{subfigure}\hfill
    \begin{subfigure}[b]{0.42\textwidth}
        \centering
        \includegraphics[width=\linewidth]{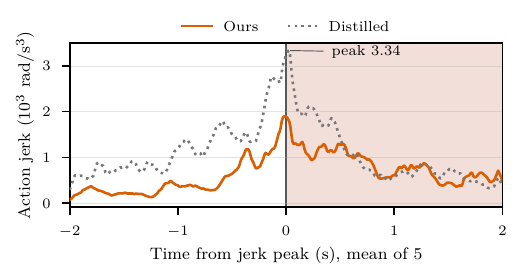}
        \vspace{-14pt}
        \caption{Hardware, aligned to the five largest jerk peaks.}
        \label{fig:hardware_jerk}
    \end{subfigure}
    \vspace{-4pt}
    \caption{Action jerk around transitions. (a) In simulation, jerk is averaged over all terrain switches of the connected course. (b) On hardware, jerk is averaged over each policy's five largest peaks, as ground-truth positions are unavailable to align transitions. The distilled policy's peak is about 1.8 times ours.}
    \label{fig:jerk}
    \vspace{-10pt}
\end{figure}

\textbf{Long-Track Results.} The connected course in Fig.~\ref{fig:long_track_summary} tests whether joint training pays off during transitions. It concatenates the six training terrains with two held-out segments, rough ground and a slope ascent/descent pair, testing composition partly outside the trained task set. Over 15 trials at $v_x=0.5$\,m/s with a pure-pursuit controller, success requires completing every segment. Our policy succeeds in 14/15 trials (93.3\%), against 11/15 (73.3\%) for the hierarchical and 8/15 (53.3\%) for the distilled policy, although only the hierarchical baseline reads the terrain index.

We measure action jerk using a third-order backward finite difference (Figs.~\ref{fig:long_track_summary} and~\ref{fig:switch_aligned_jerk}). Across the course, our policy has the lowest mean jerk at $0.78\times10^3\,\mathrm{rad/s^3}$, compared with $1.15\times10^3$ for the hierarchical policy and $1.18\times10^3\,\mathrm{rad/s^3}$ for the distilled policy. Within $\pm2$ s of a terrain switch, the means remain similar at $0.73$, $1.21$, and $1.11\times10^3\,\mathrm{rad/s^3}$, respectively. The clearest difference is the peak around a switch: $2.14\times10^3\,\mathrm{rad/s^3}$ for the hierarchical policy, $1.49\times10^3$ for the distilled policy, and $0.89\times10^3\,\mathrm{rad/s^3}$ for ours.

\begin{figure*}[b]
    \vspace{-18pt}
    \centering
    \includegraphics[width=0.8\textwidth]{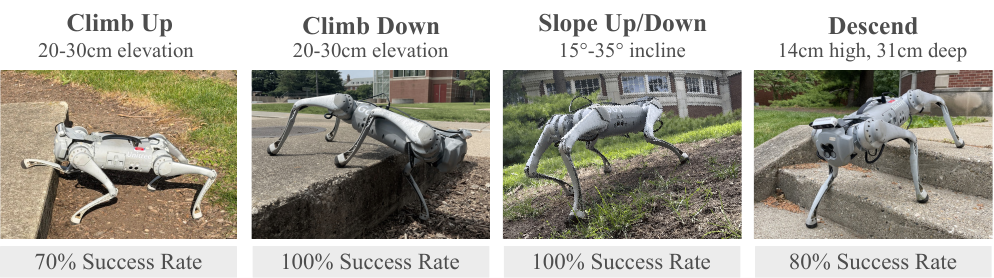}
    \caption{Real-world deployment across outdoor terrains, with success rates over 10 trials per terrain.}
    \label{fig:hardware_results}
\end{figure*}

\textbf{Hardware Results.} The policy was evaluated outdoors on terrains resembling the simulated ones, with 10 trials per terrain (Fig.~\ref{fig:hardware_results}, setup in App.~\ref{app:hardware}). Success was achieved in $100\%$ of climb-down and slope trials, $80\%$ of stair descents, and $70\%$ of climb-ups, and in all 10 trials on an extended course combining rough terrain with 12--18\,cm climb-up and climb-down obstacles. The distilled baseline was also deployed on hardware for comparison. With each policy’s five largest action-jerk peaks aligned (Fig.~\ref{fig:hardware_jerk}), the distilled policy peaks at $3.34\times10^3\,\mathrm{rad/s^3}$, against $1.90\times10^3\,\mathrm{rad/s^3}$ for ours, about 1.8 times higher. Our lower peaks suggest smoother actions even at the moments of highest action jerk, potentially reflecting smoother changes in expert contributions.

\section{Conclusion}

We studied how specialized locomotion behaviors should be composed inside a single controller. Separate critics let heterogeneous objectives be optimized jointly without suppressing specialization, and a soft router conditioned on learned perception composes the resulting behaviors continuously as terrain changes. On a connected course with held-out segments, composition yields smoother motion and avoids the switching transient of selection-based controllers, without a terrain label, selector, or distillation stage at deployment, and the reduced jerk carries over to hardware. Remaining limitations, namely weaker hardware performance on climb-up, sensitivity to close-range depth errors, and additional router hyperparameters (see App.~\ref{app:limitations}), motivate future work on uncertainty-aware routing and improved actuator modeling.


\clearpage
\acknowledgments{This work used the Delta system through allocation CIS251247 from the Advanced Cyberinfrastructure Coordination Ecosystem: Services \& Support (ACCESS) program, which is supported by National Science Foundation grants \#2138259, \#2138286, \#2138307, \#2137603, and \#2138296. This work was also supported in part by the São Paulo Research Foundation (FAPESP) under grants no. 2025/24481-5 and 2025/27983-1.}


\raggedbottom
\bibliography{example}  

\pagebreak

\appendix

\startcontents[appendix]
\begin{center}
    {\Large\bfseries Appendix}
\end{center}
\vspace{4pt}
\printcontents[appendix]{l}{1}{\setcounter{tocdepth}{2}}
\clearpage

\section{Concurrent Teacher--Student Training}
\label{app:cts}

This appendix presents the complete training objective of CTS-MoE. The background that motivates concurrent training is summarized first (App.~\ref{app:cts_background}), and the objectives optimized by each group are then given in full (App.~\ref{app:objectives}).

\subsection{Background}
\label{app:cts_background}

\textbf{Why a teacher and a student?} Perceptive locomotion is partially observable. The terrain around the feet, contact states, friction, payload, and actuator properties are unobserved or observed only through noisy, delayed sensors, but they are available exactly in simulation. A teacher that receives them solves a nearly fully observed problem and learns quickly, whereas a student restricted to deployable observations must infer them from its observation history. The teacher therefore provides a target that the student learns to reproduce, so the student is not required to discover the task from rewards under partial observability alone.

\textbf{Why concurrent rather than sequential?} In the common two-stage paradigm~\cite{kumar2021rma,agarwal2023legged}, the teacher is trained with RL and then frozen, and a student is regressed onto its latents or actions, often with DAgger-style data collection~\cite{ross2011reduction}. The actor is then optimized only on privileged latents and never experiences the errors of an imperfect student latent, the teacher may rely on information that cannot be inferred from deployable sensors, and a second training stage is required whose data coverage limits the final policy. CTS~\cite{wang2024cts} removes the second stage by splitting the parallel environments into a teacher group and a student group (85\% and 15\% here, Table~\ref{tab:hyperparameters}) that act with the same actor. Because both groups contribute PPO updates, the actor is trained continually on student-generated states and learns to act from the latent available at deployment.

\textbf{What changes in CTS-MoE?} CTS was introduced for a single task with a blind student, and four modifications are made. First, the actor is the dense MoE of Eq.~\ref{eq:moe}, so the router is trained on both latents. On the teacher group, routing can exploit the task index in $z^t_t$, whereas on the student group it must be reproduced from depth and proprioception alone, and routing errors caused by imperfect depth estimates are corrected by RL on the student group rather than inherited from a frozen policy. Second, the single critic is replaced by $N$ task-specific value heads with per-task PopArt and advantage normalization. Third, the student encoder is perceptive, combining a depth CNN and a recurrent memory with a proprioceptive-history encoder (see App.~\ref{app:encoders}). Fourth, the task index is included in the teacher latent but never given to the student.

\subsection{Optimization Objectives}
\label{app:objectives}
\label{app:training}

\textbf{PPO for two groups.} Following~\cite{wang2024cts}, each group $b\in\{t,s\}$ contributes a clipped PPO objective computed on its own data $\mathcal{D}_b$,
\begin{align}
    \mathcal{L}_{\text{PPO},t}(\theta, \theta_t) &= \frac{1}{|\mathcal{D}_t|} \sum_{\tau \in \mathcal{D}_t} \sum_{i=0}^T \min \Big( \rho_i^t \tilde{A}_i^t, \ \text{clip}(\rho_i^t, 1 - \epsilon, 1 + \epsilon) \tilde{A}_i^t \Big), \label{eq:lppo_teacher} \\
    \mathcal{L}_{\text{PPO},s}(\theta) &= \frac{1}{|\mathcal{D}_s|} \sum_{\tau \in \mathcal{D}_s} \sum_{i=0}^T \min \Big( \rho_i^s \tilde{A}_i^s, \ \text{clip}(\rho_i^s, 1 - \epsilon, 1 + \epsilon) \tilde{A}_i^s \Big), \label{eq:lppo_student}
\end{align}
where $T$ is the rollout horizon, $\epsilon$ the clipping parameter, and $\tilde{A}^b_i$ the advantage normalized per task. The probability ratios are
\begin{equation}
\rho_i^t(\theta, \theta_t) = \frac{\pi_\theta\big(a_i^t \mid \psi^t_{\theta_t}(s_i), o_i^{p}\big)}{\pi_{\theta_\text{old}}\big(a_i^t \mid \psi^t_{\theta_{t,\text{old}}}(s_i), o_i^{p}\big)}, \qquad
\rho_i^s(\theta) = \frac{\pi_\theta\big(a_i^s \mid \psi^s_{\theta_s}(o_i), o_i^{p}\big)}{\pi_{\theta_\text{old}}\big(a_i^s \mid \psi^s_{\theta_{s}}(o_i), o_i^{p}\big)}.
\label{eq:rho}
\end{equation}
The teacher ratio depends on the teacher encoder parameters, so the teacher latent is shaped by RL, whereas the student ratio depends only on the actor parameters $\theta$, since the student encoder is excluded from RL updates. Because the policy of Eq.~\ref{eq:moe} is a single Gaussian whose mean is the routed combination of expert means, both ratios are computed exactly, without a mixture likelihood.

\textbf{Per-task value learning.} Raw value targets and advantages can differ by orders of magnitude across tasks. For each task $k$, the return mean and variance are tracked by an exponential moving average and used to whiten the targets $\tilde{R}^{(k)}_t$ and predictions $\tilde{V}^{(k)}_\phi$ of Eq.~\ref{eq:value}. When these statistics change, the output layer of $V^{(k)}_\phi$ is rescaled so that its unnormalized prediction is preserved, following PopArt~\cite{van2016learning}. Advantages are likewise normalized per task before entering Eqs.~\ref{eq:lppo_teacher}--\ref{eq:lppo_student}, which prevents tasks with larger reward variance from dominating the policy gradient.

\textbf{Router regularization.} With router probabilities $p_e=\operatorname{softmax}(l)_e$ computed from logits $l$,
\begin{equation}
    \mathcal{L}_{\mathrm{ent}}=\mathbb{E}\Big[-\textstyle\sum_e p_e\log p_e\Big], \qquad
    \mathcal{L}_{\mathrm{bal}}=E\textstyle\sum_e\mathbb{E}[p_e]^2, \qquad
    \mathcal{L}_{l}=\mathbb{E}\big[\lVert l\rVert_2^2\big]
    \label{eq:moe_aux}
\end{equation}
control per-state routing entropy, encourage balanced aggregate expert use, and bound the router logits.

\textbf{Combined objective.} The actor and router $\theta$, the teacher encoder $\theta_t$, and the critics $\phi$ are updated jointly by minimizing
\begin{equation}
    \mathcal{L}(\theta,\theta_t,\phi) = -\mathcal{L}_{\text{PPO},t}-\mathcal{L}_{\text{PPO},s}+\lambda_v\mathcal{L}_{\text{value}}-\beta\,\mathcal{H}[\pi_\theta]-\lambda_{\mathrm{ent}}\mathcal{L}_{\mathrm{ent}}+\lambda_{\mathrm{bal}}\mathcal{L}_{\mathrm{bal}}+\lambda_{l}\mathcal{L}_{l},
    \label{eq:total_loss}
\end{equation}
where $\mathcal{H}[\pi_\theta]$ is the standard PPO entropy bonus and the coefficients are listed in Table~\ref{tab:hyperparameters}. The negative sign on $\mathcal{L}_{\mathrm{ent}}$ rewards router entropy and so discourages premature collapse onto a single expert. The student encoder $\theta_s$ does not appear in Eq.~\ref{eq:total_loss} and is updated only by Eq.~\ref{eq:distill}, whose stop-gradient on the teacher latent keeps the student in the same latent space without letting its errors drag the teacher toward trivial representations.
\section{Observations and Encoders}
\label{app:obs_enc}

This appendix details the observation spaces of the two groups (App.~\ref{app:observations}) and the encoders that map them to the latents of Eq.~\ref{eq:latent_states} (App.~\ref{app:encoders}).

\subsection{Observation Spaces}
\label{app:observations}

\begin{figure}[b]
    \centering
    \includegraphics[width=\textwidth]{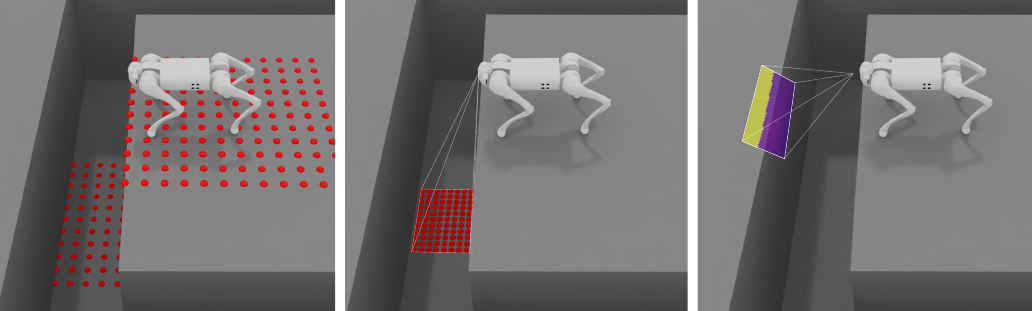}
    \caption{Perceptive observations in simulation. Left: the privileged robot-centric heightmap samples $h_t$ used by the teacher. Center: the camera rays sampling the terrain ahead of the robot. Right: the resulting depth image $d_t$, the only exteroceptive input available to the student.}
    \label{fig:sim_observations}
\end{figure}

Table~\ref{tab:observations} lists every observation term, and Fig.~\ref{fig:sim_observations} illustrates the two perceptive modalities.

\textbf{Proprioception.} The proprioceptive vector $o^p_t$ contains the velocity command, the base angular velocity and projected gravity from the IMU, joint positions and velocities from the encoders, and the previous action. All of these are measured on the Go1 and are therefore available to both groups. The base linear velocity is not reliably measured on hardware and is thus given only to the teacher; the student must infer it implicitly from the proprioceptive history $o^p_{t-H:t}$, which also allows it to identify dynamics properties such as payload or actuator response from how the robot has reacted to recent actions.

\begin{wraptable}{r}{0.44\textwidth}
    \centering
    \small
    \begin{tabular}{@{}llc@{}}
        \toprule
        \textbf{Name} & \textbf{Symbol} & \textbf{Dimension} \\ \midrule
        \multicolumn{3}{c}{\textbf{Proprioceptive} $(o^p_t)$} \\ \midrule
        Lin. Vel. Command & $v_t^{\text{cmd}}$ & 2 \\
        Ang. Vel. Command & $\omega_t^{\text{cmd}}$ & 1 \\
        Base Ang. Vel. & $\omega_t$ & 3 \\
        Proj. Gravity & $g_t$ & 3 \\
        Joint Positions & $q_t$ & 12 \\
        Joint Velocities & $\dot{q}_t$ & 12 \\
        Previous Actions & $a_{t-1}$ & 12 \\ \midrule
        \multicolumn{3}{c}{\textbf{Perceptive}} \\ \midrule
        Depth & $d_t$ &  {48 $\times$ 64} \\
        Heightmap* & $h_t$ & 187 \\ \midrule
        \multicolumn{3}{c}{\textbf{Task Embedding}} \\ \midrule
        Task ID* & $c_t$ & 1 \\ \midrule
        \multicolumn{3}{c}{\textbf{Environmental $(e_t)$}} \\ \midrule
        Base Velocity* & $v_t$ & 3 \\ 
        Contact Forces* & $F_c$ & 4 \\
        Joint Torques* & $\tau_t$ & 12 \\
        Joint Accel.* & $\ddot{q}_t$ & 12 \\
        Robot Mass* & $m$ & 1 \\
        Joint Stiffness* & $k_s$ & 12 \\
        Joint Damping* & $k_d$ & 12 \\ \bottomrule
    \end{tabular}
    \caption{Observation and privileged state spaces. Elements marked with (*) are teacher-only information.}

    \label{tab:observations}
    \vspace{-10pt}
\end{wraptable}

\textbf{Perception.} The teacher observes the terrain as a heightmap $h_t$ of 187 height samples on a grid around the robot (Fig.~\ref{fig:sim_observations}, left). This representation is exact, covers the terrain under and behind the feet, and is independent of lighting or occlusion, but it cannot be measured reliably on the robot. The student instead observes a $48\times64$ depth image $d_t$ from a forward-facing camera tilted downward (Fig.~\ref{fig:sim_observations}, center and right; App.~\ref{app:hardware}). A single depth frame shows only the terrain ahead and within the camera's valid range, so the terrain beneath the robot must be remembered from earlier frames; this is why the student encoder uses a recurrent memory (App.~\ref{app:encoders}). On hardware, depth is updated at 10\,Hz while the policy runs at 50\,Hz.

\textbf{Privileged environment parameters and task index.} The environmental vector $e_t$ collects contact forces, joint torques and accelerations, and the domain-randomized robot mass, joint stiffness, and joint damping. These make the dynamics identifiable for the teacher. The task index $c_t$ identifies which of the $N=6$ training terrains the environment belongs to. It is used during training only for selecting the reward $r^{(c_t)}$, selecting the value head $V^{(c_t)}_\phi$, and forming part of the teacher latent. It is never observed by the student or used at deployment.

\begin{figure}[b]
    \captionsetup[subfigure]{font=footnotesize}
    \centering
    \begin{subfigure}[c]{0.41\textwidth}
        \centering
        \includegraphics[width=\linewidth]{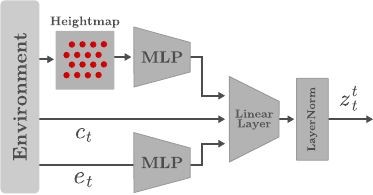}
        \caption{Teacher encoder.}
        \label{fig:teacher_encoder}
    \end{subfigure}
    \hfill
    \begin{subfigure}[c]{0.54\textwidth}
        \centering
        \includegraphics[width=\linewidth]{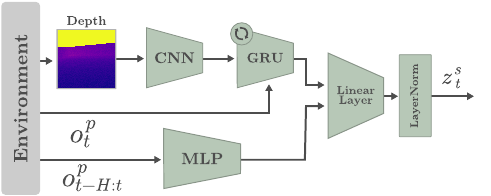}
        \caption{Student encoder.}
        \label{fig:student_encoder}
    \end{subfigure}
    \caption{Teacher and student encoder architectures.}
    \label{fig:encoders_comparison}
\end{figure}

\subsection{Encoder Architectures}
\label{app:encoders}

The teacher encodes each privileged component separately to avoid over-compressing the others. $e_t$ and $h_t$ pass through separate MLPs, whose embeddings are concatenated with the task index and projected by a linear layer with layer normalization. The student mirrors this structure with deployable inputs. An MLP over the proprioceptive history replaces the environment encoder and estimates dynamics-related information, while a CNN encodes each depth frame and a GRU fuses it with the current proprioception into a terrain memory, replacing the heightmap encoder. The task identity must be inferred by the student through the final projection. The latents are
\begin{equation}
    z_t^t = \text{LN}\big(\text{MLP}(e_t) \oplus \text{MLP}(h_t) \oplus c_t\big), \qquad
    z_t^s = \text{LN}\big(\text{MLP}(o^p_{t-H:t}) \oplus \text{GRU}(\text{CNN}(d_t) \oplus o^p_t)\big),
    \label{eq:teacher-student}
\end{equation}
where $\operatorname{LN}$ denotes a learned linear projection to 32 dimensions followed by layer normalization, which keeps the latents on a common scale so that the router and experts receive comparable inputs from either group. Figure~\ref{fig:encoders_comparison} illustrates both encoders, and Table~\ref{tab:networks} lists the layer dimensions.
\section{Simulation Setup}
\label{app:simulation}

This appendix details the simulation setup to facilitate reproducibility. All policies are trained in Isaac Lab~\cite{mittal2025isaac} with a modified RSL-RL~\cite{schwarke2025rslrl} on the Delta cluster~\cite{boerner2023access}, using four NVIDIA A40 GPUs for 15,000 iterations ($\sim$72\,h).

\textbf{Network Architectures.} Table~\ref{tab:networks} details the architectural specifications and layer dimensions for all components in our framework, ranging from the input encoders to the policy and critic networks. 

\begin{table}[h]
    \centering
    \vspace{-10pt}
    \small
    \caption{Network architecture.}
    \label{tab:networks}
    \begin{tabularx}{\linewidth}{@{} >{\raggedright\arraybackslash}X >{\raggedright\arraybackslash}X c >{\raggedright\arraybackslash}X @{}}
        \toprule
        \textbf{Component} & \textbf{Input} & \textbf{Output} & \textbf{Hidden Layers} \\ 
        \midrule
        \multicolumn{4}{c}{\textbf{Teacher Encoder} $\psi^t_{\theta_t}$} \\
        \midrule
        Privileged MLP    & $e_t$ & 32 & [512, 256] \\
        Heightmap MLP     & $h_t$ & 128 & [512, 256] \\
        Latent Projection & $c_t \oplus \mathrm{MLP}(e_t) \oplus \mathrm{MLP}(h_t)$ & 32 & Linear + LN \\
        \midrule
        \multicolumn{4}{c}{\textbf{Student Encoder} $\psi^s_{\theta_s}$} \\
        \midrule
        Proprioception MLP & $o^p_{t-H:t}$ & 32 & [512, 256] \\
        Depth CNN $\phi$   & $d_t$ & 128 & Filters [16, 32, 64] \\
        Temporal GRU       & $\phi(d_t) \oplus o^p_t$ & 256 & 1 layer \\
        Latent Projection  & $\mathrm{GRU}(\cdot) \oplus \mathrm{MLP}(o^p_{t-H:t})$ & 32 & Linear + LN \\
        \midrule
        \multicolumn{4}{c}{\textbf{Policy and Value Heads $\pi_\theta \quad V_\phi$}} \\
        \midrule
        Router MLP         & $z_t$ & $E$ & [512, 256] \\
        Actors ($\times E$) & $z_t \oplus o^p_t$ & $|\mathcal{A}|$ & [512, 256, 128] \\
        Critics ($\times N$) & $z_t \oplus o^p_t$ & 1 & [512, 256, 128] \\
        \bottomrule
    \end{tabularx}
\end{table}

\textbf{Reward Formulation.} Table~\ref{tab:rewards} lists the reward weights for each task in our multi-task reinforcement learning (MTRL) formulation. Obstacle tasks (climb up, climb down, and gap crossing) use waypoint tracking instead of velocity tracking, since projected goal points let the policy modulate its speed and gait during agile maneuvers, avoiding the rigid constraint of continuous velocity tracking~\cite{cheng2024extreme,rudin2022advanced}.

\begin{table}[htbp]
    \centering
    \vspace{-10pt}
    \caption{MTRL Reward Weights by Terrain Task.}
    \label{tab:rewards}
    \small
    
    \begin{tabularx}{\linewidth}{@{} >{\raggedright\arraybackslash}X *{6}{>{\centering\arraybackslash}X} @{}}
        \toprule
        \textbf{Reward Term} & \textbf{Flat} & \textbf{Ascend} & \textbf{Descend} & \textbf{Cross Gap} & \textbf{Climb Up} & \textbf{Climb Down} \\
        \midrule
        \multicolumn{7}{c}{\textbf{Shared Task Components}} \\
        \midrule
        $r_{v_{xy}}$        & 2.0   & 1.5   & 1.5   &   --    &    --   &   --    \\
        $r_{\omega_z}$      & 1.0   & 0.75  & 0.75  &    --   &  --     &   --    \\
        $r_{\text{wp}}$     &  --   &   --  &   --  & 1.0   & 1.0   & 1.0   \\
        \midrule
        \multicolumn{7}{c}{\textbf{Shared Regularization}} \\
        \midrule
        $r_{v_z}$           & -1.5     & -0.25    & -1.5     & -1.0     & -0.25    & -1.5     \\
        $r_{\omega_{xy}}$   & -0.05    & -0.05    & -0.05    & -0.05    & -0.05    & -0.05    \\
        $r_{q_{\text{lim}}}$& -2.0     & -2.0     & -2.0     & -2.0     & -2.0     & -2.0     \\
        $r_{\ddot q}$       & -2.5e-7  & -2.5e-7  & -2.5e-7  & -2.5e-7  & -2.5e-7  & -2.5e-7  \\
        $r_{\tau}$          & -1.0e-4  & -1.0e-4  & -1.0e-4  & -1.0e-4  & -1.0e-4  & -1.0e-4  \\
        $r_{P}$             & -2e-5    & -2e-5    & -2e-5    & -2e-5    & -2e-5    & -2e-5    \\
        $r_{\dot a}$        & -0.01    & -0.01    & -0.01    & -0.01    & -0.01    & -0.01    \\
        $r_{\ddot a}$       & -0.01    & -0.01    & -0.01    & -0.01    & -0.01    & -0.01    \\
        \midrule
        \multicolumn{7}{c}{\textbf{Task-Specific Stylistic Terms}} \\
        \midrule
        $r_{c}$             & -0.5  & -1.0  & -1.0  & -1.0  & -1.0  & -1.0  \\
        $r_{q_{\text{hip}}}$& -0.05 & -0.05 & -0.05 & -0.05 & -0.05 & -0.05 \\
        $r_{\text{feet}}$   & -0.05 & -0.2  & -0.1  & -0.05 & -0.2  & -0.1  \\
        $r_{\text{ori}}$    & -1.0  &   --  &   --  &   --  &   --  &   --  \\
        $r_{h}$             & -10.0 & -10.0 & -10.0 & -10.0 & -10.0 & -10.0 \\
        \midrule
        $r_{\text{term}}$   & -5.0  & -5.0  & -5.0  & -20.0 & -20.0 & -20.0 \\
        \bottomrule
    \end{tabularx}
\end{table}

\textbf{Hyperparameters.} We extend PPO with teacher-student groups using a latent regression process after each update, and MoE auxiliary losses controlling load balancing, router entropy, and $l$-loss regularization. Table~\ref{tab:hyperparameters} summarizes the corresponding hyperparameters.

\begin{table}[h!]
\centering
\renewcommand{\arraystretch}{1.2}
\small
\caption{Summary of training hyperparameters.}
\begin{tabular}{llc}
\toprule
\textbf{Category} & \textbf{Parameter} & \textbf{Value} \\
\midrule
\multirow{7}{*}{General PPO} 
    & Clip ratio  & 0.2 \\
    & Learning rate & $10^{-3}$ \\
    & Discount factor  & 0.99 \\
    & GAE-lambda & 0.95 \\
    & Desired KL divergence & 0.01 \\
    & Max gradient norm & 1.0 \\
    & Learning epochs / Mini-batches & 5 / 4 \\
\midrule
\multirow{2}{*}{Group Proportions of Envs} 
    & Teacher & 0.85 \\
    & Student & 0.15 \\
\midrule
\multirow{2}{*}{PPO Loss Weights}
    & Value Loss Coefficient & 1.0 \\
    & Entropy Coefficient & 0.005 \\
\midrule
\multirow{3}{*}{MoE Auxiliary Weights}
    & Balance Coefficient & $2 \times 10^{-3}$ \\
    & Router Entropy Coefficient & $5 \times 10^{-4}$ \\
    & Router $l$-Loss Coefficient & $1 \times 10^{-5}$ \\
\midrule
\multirow{1}{*}{Distillation Weights (Student)}
    & Latent regression loss & 1.0 \\
\bottomrule
\end{tabular}

\label{tab:hyperparameters}
\end{table}

\textbf{Task-Dependent Commands.} Commands $(v_x, v_y, \omega_z)$ are sampled uniformly per task. Velocity-tracking tasks use $v_x \in [-1.0, 1.0]$\,m/s, $v_y \in [-0.5, 0.5]$\,m/s, and $\omega_z \in [-0.8, 0.8]$\,rad/s. Waypoint tasks sample only forward velocities $v_x \in [0.0, 1.0]$\,m/s with $v_y=\omega_z=0$, projecting the waypoint forward.

\begin{figure}[b]
    \centering
    \includegraphics[width=\textwidth]{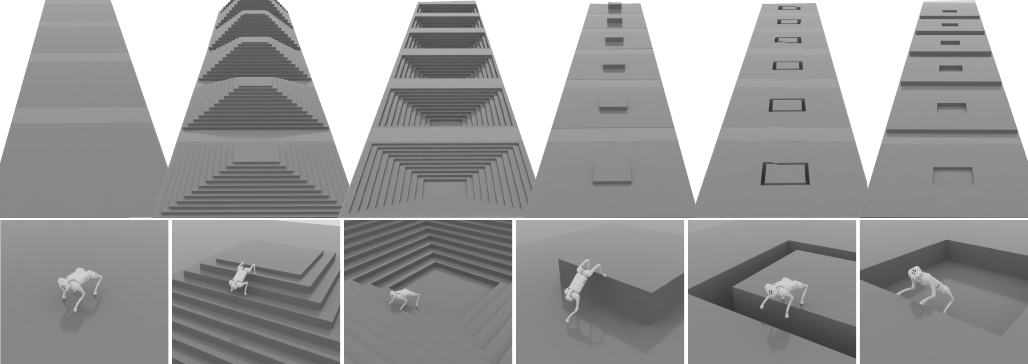}
    \caption{Training sub-terrains organized by terrain type and increasing difficulty.}
    \label{fig:curriculum}
\end{figure}

\textbf{Terrain Generation and Curriculum.} The training map comprises 36 sub-terrains arranged in a $6 \times 6$ grid, with each row corresponding to a terrain type and each column to a progressively higher difficulty level (Fig.~\ref{fig:curriculum}, Table~\ref{tab:curriculum}). Each patch spans $8 \times 8$\,m and begins with a $2 \times 2$\,m platform. We use a performance-based, probabilistic curriculum. An episode succeeds if the robot reaches the goal or traverses more than half the terrain length, and fails if it travels less than half its expected distance. To mitigate catastrophic forgetting and maintain experience diversity, curriculum transitions are stochastic. After a success, the environment advances with probability 0.8 and otherwise resets to a random easier level. After a failure, it is downgraded, with a 0.1 probability of a full reset to the base terrain. Rough, sloped, and discrete terrains are withheld from training and used to evaluate generalization.

\begin{table}[H]
    \centering
    \caption{Terrain curriculum parameters.}
    \label{tab:curriculum}
    \begin{tabular}{@{}llc@{}}
        \toprule
        \textbf{Task} & \textbf{Curriculum Parameter} & \textbf{Difficulty Range} \\
        \midrule
        \multicolumn{3}{@{}c@{}}{\textbf{Seen Terrain}} \\
        \midrule
        Flat      & ---             & --- \\
        Descend   & Step Height     & $[0.02, 0.22]$~m \\
        Ascend    & Step Height     & $[0.02, 0.22]$~m \\
        Cross Gap & Gap Width   & $[0.05, 0.50]$~m \\
        Obstacles & Obstacle Height & $[0.05, 0.50]$~m \\
        \midrule
        \multicolumn{3}{@{}c@{}}{\textbf{Unseen Terrain}} \\
        \midrule
        Rough & Frequency / Height Scale & $[18, 40]$ / $[0.12, 0.58]$~m \\
        Slope & Slope Incline            & $[0, 30]^\circ$ \\
               Discrete & Max Box Height / Intensity & $[0.05, 0.30]$~m / $[0.4, 0.7]$ \\
        \bottomrule
    \end{tabular}
\end{table}

\textbf{Domain Randomization.} Simulation results use the simulator's integrated actuators, which are efficient but poorly represent real actuators. After convergence under this setup, we resume training with extensive domain randomization, as in~\cite{he2025attention}, and an explicit delayed PD actuator model, so that the policy accounts for actuator dynamics, control latency, and non-ideal hardware. Parameters follow~\cite{wu2026robogauge}, given the similar terrain and platform. As noted in App.~\ref{app:limitations}, high-agility tasks would benefit from further techniques such as~\cite{bjelonic2025towards}.

\section{Hardware Setup}
\label{app:hardware}

\begin{wrapfigure}{r}{0.45\textwidth}
    \centering
    \vspace{-14pt}
    \includegraphics[width=\linewidth]{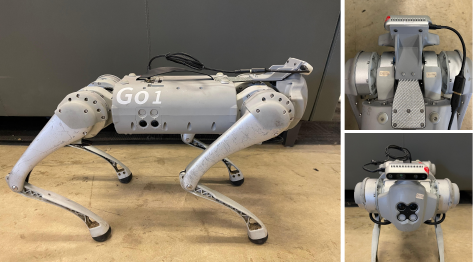}
    \caption{Unitree Go1 equipped with an Intel RealSense D435i depth camera.}
    \label{fig:real_robot}
    \vspace{-10pt}
\end{wrapfigure}

Our platform is the Unitree Go1 quadruped with 12 degrees of freedom (Fig.~\ref{fig:real_robot}). Exteroception relies on a head-mounted Intel RealSense D435i depth camera, mounted following~\cite{zhuang2023robot} and facing forward with an approximately $30^\circ$ downward tilt. Depth is downsampled to $48\times64$ pixels and updated at 10\,Hz, and valid measurements are limited to 0.2--2.0\,m. The full student policy, including depth preprocessing, CNN, and GRU inference, runs on an onboard Jetson Xavier NX at 50\,Hz. Target joint positions are passed through a second-order low-pass filter with a 12\,Hz cutoff, following~\cite{li2025reinforcement}, and tracked by a low-level PD controller ($K_p = 28$, $K_d = 0.65$) at 250\,Hz.
\section{Additional Results}
\label{app:results}

\subsection{Effect of Task-Specific Rewards and Critics}
\label{app:reward_critic}

\textbf{Full-curriculum ablation.} We change one component at a time, with every variant trained at $E=6$. Multi-reward (MR) assigns each terrain its reward from Table~\ref{tab:rewards}; shared-reward (SR) applies the flat-ground reward to all six. Holding MR fixed, multi-critic (MC) uses a separate value head and return normalization per task against shared-critic (SC) with one value head and shared statistics, so this comparison measures value separation and normalization together.

Relative to SR, MR increases curriculum progression averaged over the six training terrains by $22.6$ percentage points, driven primarily by cross-gap traversal ($+54.9$ points). With MR fixed, replacing SC with MC cuts the maximum task-return standard deviation from $1.80$ to $0.67$ ($62.8\%$) and raises effective expert utilization from $1.31$ to $5.01$, measured as the exponential of the router entropy, i.e., the equivalent number of active experts. Task-specific rewards therefore induce the behaviors each terrain requires, but only a per-task critic keeps their joint optimization stable.

Without per-task PopArt, training became unstable after approximately 750 iterations. The action standard deviation, computed across action dimensions and averaged over the rollout, grew rapidly until training collapsed.

\subsection{Per-Terrain Tracking Error}
\label{app:tracking}

Tracking quality during traversal is not captured by success alone. For an episode of length $T$, the component error is defined as
\begin{equation}
    \eta_k = (1-\mathbb{I}_{\mathrm{term}})
    \frac{1}{T}\sum_{t=1}^{T}\lVert k_t-k_t^{\mathrm{cmd}}\rVert
    +\mathbb{I}_{\mathrm{term}},\quad k\in\{v,\omega\},
\label{eq:metrics_combined}
\end{equation}
where $\mathbb{I}_{\mathrm{term}}$ indicates termination by base contact, so that a failed episode is assigned a fixed error penalty of one. Because only the linear velocity is varied across commands, $\eta=(2\eta_v+\eta_\omega)/3$ is reported, averaged over 45 episodes (15 per command $v_x\in\{0.5,0.75,1.0\}$\,m/s) for each terrain.

\begin{table*}[b]
    \vspace{-14pt}
    \centering
    \footnotesize
    \setlength{\tabcolsep}{1.1pt}
    \renewcommand{\arraystretch}{0.90}
    \newcommand{\err}[2]{\ensuremath{#1{\scriptscriptstyle\,\pm\,#2}}}
    \newcommand{\berr}[2]{\ensuremath{\mathbf{#1}{\scriptscriptstyle\,\pm\,#2}}}
    \newcommand{\uerr}[2]{\ensuremath{\underline{#1}{\scriptscriptstyle\,\pm\,#2}}}
    \caption{Velocity-tracking error $\eta$ by curriculum-level band and terrain. Lower is better.}
    \label{tab:metrics}
    \begin{threeparttable}
        \begin{tabularx}{\textwidth}{@{}p{0.125\textwidth}c*{8}{>{\centering\arraybackslash}X}@{}}
            \toprule
            \textbf{Method} & \textbf{Diff.} & \textbf{F} & \textbf{A} & \textbf{D} & \textbf{G} & \textbf{CU} & \textbf{CD} &\textbf{DT$^*$} & \textbf{R$^*$} \\
            \midrule
            \multirow{3}{*}{\textbf{Ours}}
            & 0--1 & \berr{0.09}{0.05} & \berr{0.08}{0.02} & \berr{0.07}{0.02} & \uerr{0.26}{0.10} & \err{0.26}{0.03} & \err{0.18}{0.04} & \berr{0.07}{0.01} & \berr{0.10}{0.01} \\
            & 2--3 & \berr{0.10}{0.07} & \berr{0.09}{0.02} & \berr{0.08}{0.02} & \berr{0.28}{0.04} & \err{0.26}{0.08} & \err{0.20}{0.07} & \berr{0.08}{0.01} & \berr{0.10}{0.01} \\
            & 4--5 & \err{0.15}{0.08} & \uerr{0.20}{0.25} & \berr{0.11}{0.02} & \berr{0.32}{0.16} & \uerr{0.57}{0.37} & \err{0.32}{0.15} & \berr{0.07}{0.01} & \berr{0.11}{0.02} \\
            \midrule
            \multirow{3}{*}{\begin{tabular}[c]{@{}l@{}}\textbf{Perceptive-}\\\textbf{Single}\end{tabular}}
            & 0--1 & \uerr{0.10}{0.03} & \uerr{0.11}{0.03} & \uerr{0.08}{0.02} & \berr{0.25}{0.24} & \err{0.27}{0.27} & \uerr{0.15}{0.05} & \uerr{0.11}{0.04} & \uerr{0.14}{0.04} \\
            & 2--3 & \uerr{0.12}{0.04} & \uerr{0.12}{0.02} & \uerr{0.09}{0.02} & \err{0.39}{0.36} & \uerr{0.25}{0.20} & \uerr{0.19}{0.03} & \err{0.16}{0.05} & \uerr{0.15}{0.05} \\
            & 4--5 & \berr{0.13}{0.08} & \err{0.32}{0.34} & \uerr{0.13}{0.03} & \err{0.66}{0.39} & \err{0.65}{0.38} & \err{0.30}{0.25} & \err{0.16}{0.05} & \err{0.15}{0.04} \\
            \midrule
            \multirow{3}{*}{\textbf{Blind-MoE}}
            & 0--1 & \err{0.12}{0.03} & \err{0.18}{0.04} & \err{0.11}{0.02} & \err{0.34}{0.34} & \err{0.23}{0.18} & \berr{0.13}{0.03} & \err{0.12}{0.02} & \err{0.16}{0.04} \\
            & 2--3 & \err{0.14}{0.04} & \err{0.28}{0.27} & \err{0.13}{0.03} & \err{0.72}{0.40} & \err{0.37}{0.31} & \berr{0.16}{0.10} & \uerr{0.12}{0.02} & \err{0.19}{0.04} \\
            & 4--5 & \uerr{0.14}{0.04} & \err{0.79}{0.36} & \err{0.21}{0.09} & \err{0.94}{0.21} & \err{0.91}{0.24} & \uerr{0.28}{0.26} & \uerr{0.11}{0.02} & \err{0.17}{0.04} \\
            \midrule
            \multirow{3}{*}{\textbf{Blind-Single}}
            & 0--1 & \err{0.22}{0.04} & \err{0.17}{0.10} & \err{0.18}{0.04} & \err{0.30}{0.24} & \berr{0.18}{0.10} & \err{0.20}{0.04} & \err{0.21}{0.04} & \err{0.23}{0.05} \\
            & 2--3 & \err{0.22}{0.07} & \err{0.16}{0.16} & \err{0.17}{0.03} & \err{0.78}{0.34} & \berr{0.24}{0.19} & \err{0.20}{0.04} & \err{0.20}{0.04} & \err{0.18}{0.05} \\
            & 4--5 & \err{0.22}{0.05} & \err{0.50}{0.42} & \err{0.24}{0.05} & \err{0.99}{0.06} & \err{0.71}{0.38} & \err{0.29}{0.18} & \err{0.19}{0.05} & \uerr{0.13}{0.02} \\
            \midrule
            \multirow{3}{*}{\begin{tabular}[c]{@{}l@{}}\textbf{Hierarchical}\\\textbf{(Oracle)}\end{tabular}}
            & 0--1 & \err{0.07}{0.02} & \err{0.08}{0.02} & \err{0.07}{0.04} & \err{0.27}{0.27} & \err{0.23}{0.01} & \err{0.21}{0.02} & \err{0.18}{0.02} & \err{0.19}{0.02} \\
            & 2--3 & \err{0.12}{0.02} & \err{0.12}{0.05} & \err{0.13}{0.05} & \err{0.34}{0.24} & \err{0.40}{0.34} & \err{0.22}{0.15} & \err{0.27}{0.22} & \err{0.22}{0.03} \\
            & 4--5 & \err{0.18}{0.03} & \err{0.18}{0.06} & \err{0.17}{0.08} & \err{0.37}{0.36} & \err{0.60}{0.40} & \err{0.29}{0.27} & \err{0.24}{0.01} & \err{0.74}{0.37} \\
            \midrule
            \multirow{3}{*}{\textbf{Distilled}}
            & 0--1 & \err{0.18}{0.05} & \err{0.13}{0.04} & \err{0.13}{0.06} & \err{0.33}{0.38} & \uerr{0.19}{0.04} & \err{0.19}{0.03} & \err{0.41}{0.20} & \err{0.48}{0.25} \\
            & 2--3 & \err{0.17}{0.05} & \err{0.14}{0.08} & \err{0.11}{0.04} & \uerr{0.37}{0.36} & \err{0.40}{0.35} & \err{0.29}{0.27} & \err{0.58}{0.36} & \err{0.68}{0.27} \\
            & 4--5 & \err{0.22}{0.10} & \berr{0.17}{0.08} & \err{0.17}{0.11} & \uerr{0.61}{0.39} & \berr{0.41}{0.32} & \berr{0.22}{0.31} & \err{0.76}{0.34} & \err{0.71}{0.29} \\
            \bottomrule
        \end{tabularx}
        \begin{tablenotes}[para,flushleft]
            \small Mean $\pm$ standard deviation. Bold and underline mark the lowest and second-lowest error among deployable policies (the Hierarchical oracle is excluded). F is flat, A/D stairs, G cross gap, CU/CD climb, DT discrete terrain, and R rough. $^*$Unseen terrain.
        \end{tablenotes}
    \end{threeparttable}
\end{table*}

As shown in Table~\ref{tab:metrics}, the lowest tracking error among deployable policies is achieved by ours in 15 of the 24 terrain--difficulty cells, including every band of the held-out discrete and rough terrains. The remaining cells are led by Perceptive-Single on gaps at the easiest band and on flat ground at the hardest band, by Blind-Single on climb-up and Blind-MoE on climb-down at the two easier bands, and by the distilled policy on ascent, climb-up, and climb-down at the hardest band. Success and tracking therefore diverge at high difficulty, since our policy stays upright where the others terminate but does not always track most closely once committed to a traverse.

The two specialist-based baselines behave differently on seen and unseen terrain. The hierarchical oracle matches or improves on our error at the easiest band of flat ground, ascent, and descent, where each specialist operates on the terrain it was trained for. On the held-out terrains, however, both degrade sharply, with errors of $0.18$--$0.27$ (hierarchical) and $0.41$--$0.76$ (distilled) on discrete terrain and up to $0.74$ and $0.71$ on rough terrain, against at most $0.11$ for ours. Since every method succeeds on rough terrain, this gap reflects tracking quality rather than failures, and it is consistent with better generalization through composition than through fixed specialists.

\subsection{Terrain-Specific Expert Usage}
\label{app:experts}

\begin{figure}[b]
    \centering
    \begin{subfigure}[b]{0.48\textwidth}
        \centering
        \includegraphics[width=\linewidth]{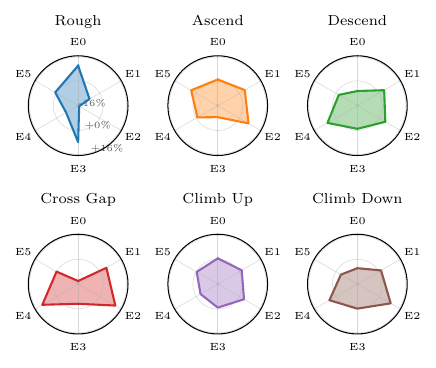}
        \caption{Terrain-relative expert usage.}
        \label{fig:expert_patterns}
    \end{subfigure}\hfill
    \begin{subfigure}[b]{0.48\textwidth}
        \centering
        \includegraphics[width=\linewidth]{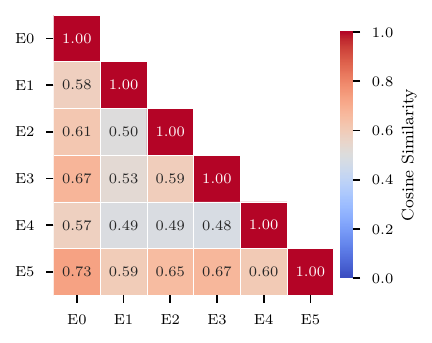}
        \caption{Pairwise expert cosine similarity.}
        \label{fig:expert_cosine}
    \end{subfigure}
    \caption{Expert usage and similarity of the student policy. (a) Average routing weight of each expert on each terrain, relative to its mean usage across terrains. (b) Pairwise cosine similarity between experts.}
    \label{fig:experts}
\end{figure}

Figure~\ref{fig:expert_patterns} shows each expert's average routing weight on each terrain relative to its mean usage across terrains. Each terrain exhibits a distinct but overlapping pattern, suggesting that terrain-dependent behavior emerges through compositions of shared experts rather than a one-to-one expert-to-terrain assignment. These compositions are learned without explicit task identification at deployment or a prescribed expert-to-task assignment. The connected-course router trace in Fig.~\ref{fig:long_track_summary} shows a corresponding temporal structure, with several experts active simultaneously, rapid weight shifts near terrain boundaries, and more stable patterns within homogeneous segments.

Figure~\ref{fig:expert_cosine} complements this view with the pairwise cosine similarity between the experts' mean actions $\mu_e$. Off-diagonal similarities range from $0.48$ to $0.73$, indicating shared action structure alongside substantial differences between experts. Expert $E_4$ has the lowest average similarity to the others, whereas $E_0$, $E_3$, and $E_5$ form a relatively similar group ($0.67$--$0.73$). Together with the routing patterns, these results are consistent with a shared locomotion structure and terrain-dependent expert composition.

Replacing the router outputs with uniform weights $g_e=1/E$ at inference, while keeping all other parameters fixed, reduces overall success across terrains to $2.2\%$. Despite the experts' shared action structure, successful locomotion therefore depends on their learned, observation-dependent weighting.

\subsection{Number of Experts}
\label{app:num_experts}

The number of experts is the main additional design choice of the mixture and is the source of the hyperparameter limitation noted in App.~\ref{app:limitations}. Since $E$ need not equal the number of tasks, variants with $E=4$ and $E=2$ were also trained. At $E=4$, $89.4\%$ of the $E=6$ success rate was retained, but only after the router coefficients were retuned. At $E=2$, no tested coefficient setting exceeded $22.4\%$ of the $E=6$ success rate, below the single-actor baseline, which indicates that the routing regularizers of Eq.~\ref{eq:moe_aux} become counterproductive when experts are too few to specialize. As $E$ decreases, these terms become harder to balance, so more hyperparameter tuning is required than for single-actor alternatives.

\newpage
\subsection{Training Curves and Curriculum Progression}
\label{app:training_metrics}

\begin{figure}[t]
    \centering
    \begin{subfigure}[b]{0.9\textwidth}
        \centering
        \includegraphics[width=\linewidth]{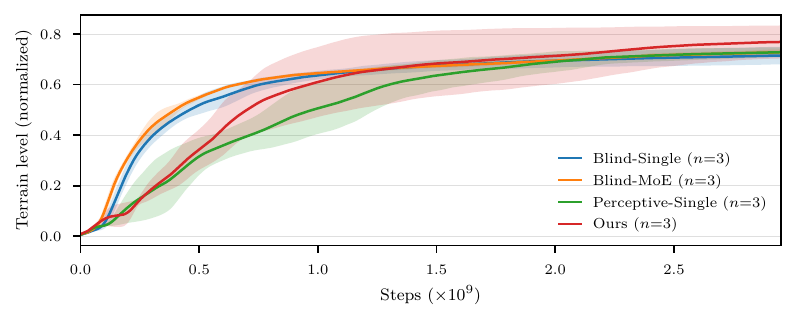}
        \vspace{-16pt}
        \caption{Normalized terrain level, mean over three seeds with the spread across seeds shaded.}
        \label{fig:level_curves}
    \end{subfigure}
    \\[4pt]
    \begin{subfigure}[b]{0.9\textwidth}
        \centering
        \includegraphics[width=\linewidth]{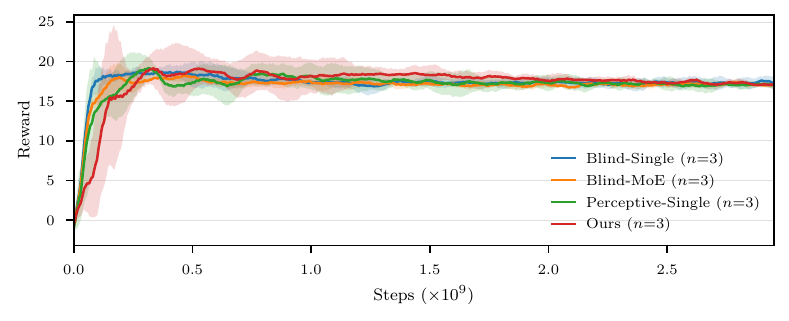}
        \vspace{-16pt}
        \caption{Average reward.}
        \label{fig:reward_curves}
    \end{subfigure}
    \caption{Training progress over simulation steps.}
    \label{fig:rewards}
\end{figure}

Reward alone provides limited insight into training progress, since it converges quickly while similar rewards correspond to progressively harder terrain as the curriculum advances (Fig.~\ref{fig:reward_curves}), so curriculum progression is used instead (Fig.~\ref{fig:level_curves}). All methods were trained with three seeds, and the plotted curves show limited variation in final curriculum levels across seeds. Evaluation results are reported for the best seed of each method, selected consistently by final terrain level. The blind policies advance fastest early in training, whereas the perceptive policies advance more slowly, since the depth encoder and recurrent memory must be learned before perception becomes useful. Our policy then overtakes the blind policies and reaches the highest final level, while Perceptive-Single catches up only late in training. Blind-MoE ends close to Blind-Single, so the mixture benefits progression mainly when routing is conditioned on perception, consistent with Sec.~\ref{sec:experiments}.

\section{Limitations}
\label{app:limitations}

Hardware performance is weakest on climb-up, where higher joint reactivity amplifies residual actuator mismatch, and gaits remained less regular than in simulation despite dynamics randomization and action filtering. Improved actuator modeling, such as the identification approach of~\cite{bjelonic2025towards}, would likely reduce this gap. Since routing is inferred from depth, valid only between 0.2 and 2.0\,m, close-range sensing errors and outdoor lighting can produce inappropriate expert mixtures and brief jerky corrections, which uncertainty-aware routing could mitigate. The mixture also adds hyperparameters absent from single-actor policies, as the number of experts and the router regularizers must be tuned together (see App.~\ref{app:num_experts}).

\end{document}